# Hyperedge Interaction-aware Hypergraph Neural Network

Rongping Ye, Xiaobing Pei, Haoran Yang, Ruiqi Wang
Huazhong University of Science and Technology
m202276628@hust.edu.cn, xiaobingp@hust.edu.cn, m202276630@hust.edu.cn, m202376887@hust.edu.cn.

## ABSTRACT

Hypergraphs provide an effective modeling approach for modeling high-order relationships in many real-world datasets. To capture such complex relationships, several hypergraph neural networks have been proposed for learning hypergraph structure, which propagate information from nodes to hyperedges and then from hyperedges back to nodes. However, most existing methods focus on information propagation between hyperedges and nodes, neglecting the interactions among hyperedges themselves. In this paper, we propose HeIHNN, a hyperedge interaction-aware hypergraph neural network, which captures the interactions among hyperedges during the convolution process and introduce a novel mechanism to enhance information flow between hyperedges and nodes. Specifically, HeIHNN integrates the interactions between hyperedges into the hypergraph convolution by constructing a three-stage information propagation process. After propagating information from nodes to hyperedges, we introduce a hyperedge-level convolution to update the hyperedge embeddings. Finally, the embeddings that capture rich information from the interaction among hyperedges will be utilized to update the node embeddings. Additionally, we introduce a hyperedge outlier removal mechanism in the information propagation stages between nodes and hyperedges, which dynamically adjusts the hypergraph structure using the learned embeddings, effectively removing outliers. Extensive experiments conducted on real-world datasets show the competitive performance of HeIHNN compared with state-of-the-art methods.

## 1 INTRODUCTION

The Graph structures are commonly found in various application scenarios [1, 2, 24], and the use of graph neural networks has gained increasing attention due to their ability to leverage the power of graph structures. Graphs can model pairwise relationships between entities, but they are unable to capture high-order relationships among multiple entities [3, 4, 5, 6]. As a generalized structure of graphs, hypergraphs define hyperedge, as a collection of nodes, that can connect more than two nodes. Therefore, extending graph neural networks (GNNs) to hypergraphs enables the handling of multi-modal data and the capture of high-order correlations in the data [23, 27].

However, the application of Graph Neural Networks (GNNs) to hypergraphs presents a challenge due to the high-order relationships among multiple nodes represented by hyperedges. To address this, one must either preprocess the hypergraph structure to obtain a format compatible with graph convolution or define convolution operations directly on the hypergraph [15, 28]. It requires preprocessing the hyper-graph structure to obtain a format that is compatible with graph convolution or defining convolution operations directly on the hypergraph [27, 28]. One straightforward approach is to transform the hypergraph into a simple graph that accurately represents the hypergraph while retaining as much information as possible [7, 8, 9, 12]. This simple graph only contains pairwise relationships, enabling the application of a graph neural network (GNN). Alternatively, convolution operations can be directly defined on the hypergraph structure without converting it into a simple graph [17, 18, 19, 20]. This method preserves the comprehensive semantic relationships between nodes and the higher-order relationships within the hypergraph, avoiding loss of information [20].

Although the above methods have achieved promising results, they primarily focus on information propagation between nodes and hyperedges, neglecting the interaction among hyperedges. However, the interaction between hyperedges can effectively model the high-order interactions among diverse entities/nodes that widely exist in real-world scenarios. Moreover, the hyperedge features, as representations of these sets of nodes, can effectively indicate the collective characteristics of the nodes within them, which can be further utilized for subsequent operations and downstream tasks. Therefore, it is valuable to incorporate hyperedge interactions into hypergraph convolution to capture rich hyperedge information.

In this paper, we propose a **H**yp**e**redge **I**nteraction-aware **H**ypergraph **N**eural **N**etwork (HeIHNN) model, to learn hypergraph representation. Specifically, our approach integrates the interactions between hyperedges into the hypergraph convolution process by designing a three-stage hypergraph information propagation process: node-to-hyperedge (N2HE), hyperedge-to-hyperedge (HE2HE), and hyperedge-to-node (HE2N). Firstly, we aggregate the information from all nodes within each hyperedge to update the embedding of the hyperedge. Next, we construct a hyperedge interaction graph based on the relationships between hyperedges, enabling the convolutional propagation of information at the hyperedge level. Lastly, we update the embeddings of all nodes within the hyperedges using the acquired hyperedge features. For each stage, we design corresponding convolutions and incorporate attention mechanisms to capture the significance of different components. Additionally, we propose a hyperedge outlier removal mechanism that dynamically adjusts the hypergraph structure by identifying and removing outliers within the hyperedges during the information propagation between nodes and hyperedges. To evaluate the performance of HeIHNN, we conduct experiments on five real-world datasets and compare our results with existing models. The experimental results demonstrate that HeIHNN achieves competitive performance by effectively incorporating information propagation between hyperedges and optimizing the interaction between nodes and hyperedges.

Our major contributions are as follows:
- We propose HeIHNN, a novel hypergraph neural network model that integrates the interactions between hyperedges into the hypergraph convolution for representation learning of hypergraph structures. HeIHNN effectively handles high-order data and models relationships among hyperedges.
- We introduce a hyperedge outlier removal mechanism during the information propagation process between nodes and hyperedges. This mechanism dynamically adjusts the hypergraph structure using learned embeddings, effectively removing outliers and enhancing the ability of model to handle noise and irrelevant information.
- We conduct extensive experiments of node classification on five real-world datasets to validate the performance of HeIHNN.

## 2   RELATED WORK

In recent years, graph neural networks (GNNs) have made rapid progress and become increasingly refined on simple graphs. People have started considering more complex graph structures such as heterogeneous graphs [25, 32], dynamic graphs [26, 35], and hypergraphs [7, 12]. In the real world, relationships often involve interactions among two or more nodes, rather than pairwise connections. As a result, there has been a growing interest in extending GNNs to hypergraphs [23].

Hypergraph neural networks with expansion transform the hypergraph into a corresponding simple graph and then utilize GNNs to learn the graph structure. HGNN, proposed by Feng et al., defines the hypergraph Laplacian operator and constructs the hypergraph convolution process, enabling the migration of graph neural networks to hypergraphs [7, 14]. DHGNN, proposed by Jiang et al., addresses the possibility of imperfect initial hypergraph structures by reconstructing the hypergraph using updated node features after each convolution [8]. HGC-RNN, proposed by Yi and Park, extends HGNN by incorporating an RNN to learn temporal features of sequential hypergraphs [9]. Chen et al. propose a multi-level graph neural network that can handles both paired graphs and hypergraphs [10]. Bai et al. further incorporates an attention mechanism to HGNN, which measures the weights of different nodes during the aggregation process [11]. Yadati et al. point out that the dense subgraphs generated by clique expansion can lead to increased complexity and propose HyperGCN, a simplified version of clique expansion [12]. Yang et al. propose linear expansion, which creates a new type of node for each node and the hyperedge containing it, called a line node. Two types of edges are constructed based on shared nodes or edges between line nodes, and separate information propagation methods are designed for each type of edge [15].

To preserve the original semantic information as much as possible, hypergraph neural networks without expansion directly define convolution operations on the hypergraph structure. Arya et al. drew inspiration from spatial methods like GraphSage [16] and proposed HyperSage [17], which employs a two-level aggregation process: first aggregating node features within the same hyperedge to generate hyperedge features, and then aggregating hyperedge features containing the central node to obtain the embedding representation of the central node. HyperGAT, proposed by Ding et al., further incorporates an attention mechanism to measure the importance of different features during the aggregation process [18]. Hyper-SAGNN, proposed by Zhang et al., is applicable to hyperedges of arbitrary sizes and heterogeneity [19]. HNHN, proposed by Dong et al., apply nonlinear activation functions to both hypernodes and hyperedges and incorporates a normalization scheme that can flexibly adjust the importance of high-cardinality hyperedges and nodes [36]. AllSet, proposed by Chien et al., integrates Deep Sets and Set Transformers with hypergraph neural networks to learn two multiset functions which represent hypergraph neural network layers, allowing for greater modeling flexibility and expressive power [20].

## 3   PRELIMINARIES

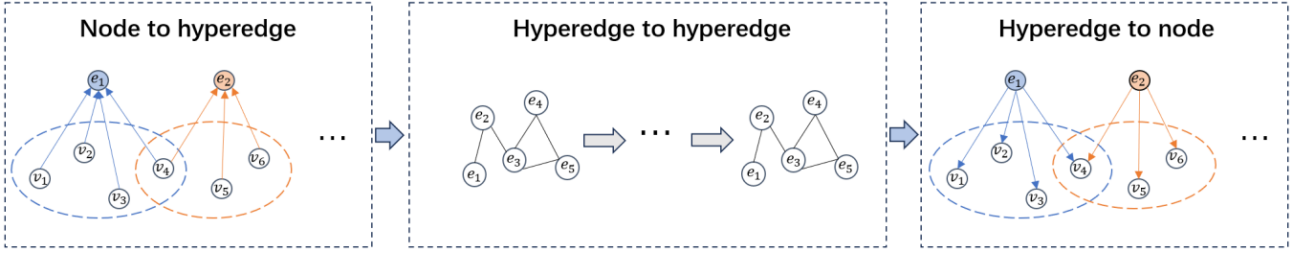

**Figure 1: The illustration of the 3-stage hypergraph information propagation process.**

A hypergraph is an extension of the definition of a simple graph, specifically in terms of edge. In a hypergraph, the edges are called hyperedges, and a hyperedge can contain an arbitrary number of nodes. This allows for the direct representation of higher-order relationships and provides a more powerful expressive capability for complex relationships.

Similar to a simple graph, a hypergraph $G = (V, E)$ is represented, where $V = \{v_1, v_2, ..., v_n\}$ is the set of nodes and $E = \{e_1, e_2, ..., e_m\}$ is the set of hyperedges. Each hyperedge $e$ in $E$ is a collection of multiple nodes. The incidence matrix $H \in R^{|V|*|E|}$ is defined such that $H(v, e) = 1$ if and only if $v \in e$; otherwise, $H(v, e) = 0$. The diagonal matrices $D_V$ and $D_E$ are the degree matrices for nodes and hyperedges, respectively. The relevant notations and definitions are shown in Table 1.

**Table 1: Notations and definitions.**

| Notation | Definition |
|---|---|
| $G$ | $G = (V, E, H)$ indicates a hypergraph, $V$ and $E$ indicates the set of nodes and the set of hyperedges, respectively. |
| $V$ | The set of nodes of $G$. |
| $E$ | The set of hyperedges of $G$. |
| $H$ | The incidence matrix, $H(v, e) = 1$ if and only if $v \in e$; otherwise, $H(v, e) = 0$. |
| $n$ | The number of nodes on $G$. |
| $m$ | The number of hyperedges on $G$. |
| $x_i^0$ | The initial feature for the $i$-th node on $G$. |
| $y_j^0$ | The initial feature for the $j$-th hyperedge on $G$. |
| $x_i$ | The representation of node $v_i$. |
| $y_i$ | The representation of hyperedge $e_i$. |
| $X^0$ | The initial feature for all the nodes on $G$. |
| $Y^0$ | The initial feature for all the hyperedges on $G$. |
| $X^l$ | The node embedding in layer $l$. |
| $Y^l$ | The hyperedge embedding in layer $l$. |
| $D_V$ | The diagonal matrices of node degrees. |
| $D_E$ | The diagonal matrices of hyperedge degrees. |

## 4 METHOD

In this section, we will introduce our proposed hypergraph neural network model HeIHNN that integrates the interactions between hyperedges into the hypergraph convolution by constructing a three-stage information propagation process. We also introduce a hyperedge outlier removal mechanism that is applied to the information propagation between hyperedges and nodes.

### 4.1 3-Stage Information Propagation Process

The three-stage information propagation process of HeIHNN we propose is illustrated in Figure 1: the circles represent hyperedges, and the nodes within them represent the inclusion relationships of the hyperedges. In the first stage, the features of nodes belonging to each hyperedge are aggregated to update the corresponding hyperedge features. In the second stage, connections are established between hyperedges that share common nodes (For instance, $v_4$ is simultaneously included in hyperedges $e_1$ and $e_2$), resulting in the construction of the hyperedge interaction graph $G_{he}$. This graph structure is learned and utilized to update the hyperedge features. In the third stage, the updated hyperedge features are employed to update the features of the nodes within them.

*4.1.1 Node to Hyperedge Stage (N2HE).* During the initial stage of information propagation, the information from all nodes within a hyperedge is aggregated to update the hyperedge embedding. One common approach is to compute the weighted average of the node features within the hyperedge. However, it is crucial to consider that different nodes within the hyperedge may have varying levels of influence on the overall hyperedge representation. Nodes that exert a stronger impact on the hyperedge should exhibit higher similarity with the hyperedge's features. Given a hyperedge $e_i$, we update the hyperedge embedding by aggregating the information of all nodes on the hyperedge and weighting it with the existing hyperedge features:

$$y_i = \sum_{v_j \in N_{e_i}} \alpha_{ij} Mess(x_j) + y_i \quad (1)$$

here, $N_{e_i}$ represents the set of nodes on hyperedge $e_i$, $Mess_{v \to e}(\cdot)$ is the information generation function from nodes to hyperedges, $\alpha_{ij}$ is the weight.

*4.1.2 Hyperedge to Hyperedge Stage (HE2HE).* Next, we propagate information between hyperedges. If two hyperedges share the same nodes, we consider there to be an interaction between these hyperedges. Based on above assumption, we can treat hyperedges as he-nodes and connect two he-nodes if their intersection is not empty. Hence, we construct a hyperedge interaction graph $G_{he} = (A_{he}, Y)$ to represent the relationships between hyperedges. Here, $A_{he} \in R^{|E|*|E|}$ is the adjacency matrix of the hyperedge interaction graph, $Y \in R^{|E|*d}$ is the feature matrix of the hyperedges, $d$ is the feature dim. The adjacency matrix $A_{he}$ can be derived from the incidence matrix $H$ as follows [13, 21, 22]:

$$A_{he} = H^T H \quad (2)$$

if the non-zero element $a_{ij} = n$ in the adjacency matrix $A_{he}$, where $n$ is a positive integer, it implies that hyperedge $e_i$ and hyperedge $e_j$ share $n$ nodes in the hypergraph. So, the elements in $A_{he}$ can be directly used as the weights for information propagation between hyperedges.

Afterwards, we apply GNN to learn $G_{he}$ and update the hyperedge embeddings. The update can be represented as:

$$Y' = GNN(A_{he}, Y) \quad (3)$$

If we only use a single layer of GNN, the hyperedges can only aggregate information from their first-order neighbors before propagating it to the nodes. To gather information from high-order neighborhoods, we can incorporate the k-order Chebyshev formula:

$$Y' = \sigma\left(U\left(\sum_{i=0}^{K}\beta_i T_i(\Lambda')\right) U^T Y \theta_2^{(l)}\right) \quad (4)$$

here $\Lambda' = \frac{\Lambda}{\lambda_{\max}} - I_m$, $\Lambda$ is the diagonal matrix of eigenvalues corresponding to the Laplacian matrix of $A_{he}$, and $U$ is the matrix of eigenvectors.

*4.1.3 Hyperedge to Node Stage (HE2N).* After completing the information propagation between hyperedges, the hyperedges need to propagate information to all nodes within them. From the perspective of a node $v_i$, this is equivalent to aggregating information from all hyperedges that contain $v_i$. Given a node $v_i$, we update the node embedding by aggregating information from all hyperedges that contain this node, and weighting it with the existing node features:

$$x_i = \sum_{e_j \in N_{v_i}} \beta_{ij} Mess(y_j) + x_i \quad (5)$$

here $N_v$ represents the set of hyperedges that contain node $v_i$, $Mess_{e\to v}(\cdot)$ denotes the information generation function from hyperedges to nodes, and $\beta_{ij}$ represents the weight.

## 4.2 Hyperedge Outlier Removal Mechanism

During the construction of a hypergraph, existing simple graph structures and K-nearest neighbors (KNN) algorithms based on different similarity metrics are often utilized. However, certain edges in the existing graph structure may represent weak or incorrect connections. In the case of the KNN method, the value of k, which determines the number of nodes included in each hyperedge, needs to be predetermined. When the k value is large for specific hyperedges, it can lead to the inclusion of irrelevant nodes within those hyperedges. It is crucial to recognize that existing hypergraph structures may not always be suitable [14].

To address the issue of including inappropriate nodes within hyperedges, we propose a **H**yperedge **O**utlier **R**emoval mechanism (HOR), as illustrated in Figure 2. The circle represents hyperedges $e_i$, and the nodes $v_{ij}$ are the nodes contained within the hyperedge. The proximity of a node to the center of a circle represents the similarity between the node and the hyperedge.

Since the features of a hyperedge represent the collective characteristics of all nodes within that hyperedge, nodes that exerting a significant influence should exhibit higher similarity to the hyperedge features. Given hyperedge $e_j$ and its constituent node $v_i$, we employ cosine similarity to calculate the similarity between the node's feature $x_i$ and the hyperedge feature $y_j$:

$$cos_{ij} = \frac{x_i \cdot y_j}{|x_i| \cdot |y_j|} \quad (6)$$

where $|\cdot|$ is the norm operation on vectors. Nodes that exhibit lower similarity to the hyperedge features are considered to have a weak influence on the hyperedge.

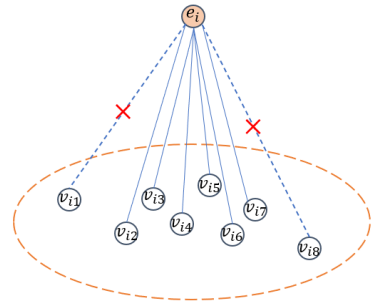

**Figure 2: Hyperedge outlier removal mechanism applied on a hyperedge.**

Therefore, we leverage the similarity between each node and its corresponding hyperedge to identify and remove outlier nodes. Considering the similarity values, we can identify nodes that have a relatively small influence on the hyperedge. These outlier nodes are subsequently removed from the hyperedge, resulting in an optimized hypergraph structure with reduced complexity. This process enhances the coherence and quality of the hypergraph, leading to improved performance of HeIHNN. HOR mechanism can be applied during both N2HE and HE2N stages. During the learning process, incorporating HOR mechanism allows for dynamic optimization of the hypergraph structure using the learned embeddings.

## 4.3 Hypergraph Convolution Layer

The hypergraph incidence matrix $H$ and the node feature $X$ are fed into the model to get the labels. According to the proposed information propagation process, we construct the formal expression of the $l-th$ layer of HeIHNN convolution. To begin with, the aggregation of node information within a hyperedge can be represented as $H^T X^{(l)}$, the method incorporating standardization is $D_e^{-\frac{1}{2}} H^T D_v^{-\frac{1}{2}} X^{(l)}$. Then add the node information to the existing information within the hyperedge to update hyperedge embeddings and introducing a hyperparameter $\alpha$ to control the impact of the node information, which can be expressed as follows:

$$Y^{(l,1)} = \alpha D_e^{-\frac{1}{2}} H^T D_v^{-\frac{1}{2}} X^{(l)} + Y^{(l)} \quad (7)$$

Introducing self-attention mechanism further enables the automatic learning of weight coefficients for each node in the information aggregation step:

$$e_{ij} = (W_Q^T x_i)^T (W_K^T y_j) \quad (8)$$

$$\alpha_{ij} = \frac{\exp(e_{ij})}{\sum_{v_k \in N_{e_i}} \exp(e_{kj})} \quad (9)$$

where $W_Q$ and $W_K$ are parameter matrices. The computed attention coefficients between node $v_i$ and hyperedge $e_j$ are assigned to the corresponding element of the hypergraph incidence matrix $H$, resulting in an attention incidence matrix $H^{Att} = \{h_{ij}^{Att} = \alpha_{ij}, if\ h_{ij} \neq 0, h_{ij} \in H\}$. The output of N2HE stage can be expressed as:

$$Y^{(l,1)} = \sigma \left( \left( \alpha_1 D_e^{-\frac{1}{2}} (hor(H^{Att}))^T D_v^{-\frac{1}{2}} X^{(l)} + Y^{(l)} \right) \theta_1^{(l)} \right) \quad (10)$$

where $Y^{(l)}$ is the $l-th$ layer hyperedge embeddings, $X^{(l)}$ is the $l-th$ layer node embeddings, $\sigma$ denotes the non-linear activation function, $\theta_1^{(l)}$ is a learnable parameter matrix, $hor(\cdot)$ is the hyperedge outlier removal function.

The output $Y^{(l,1)}$ from the N2HE stage is fed as the updated features of the hyperedges into the HE2HE stage. For the information propagation between hyperedges, since the constructed HE-graph is a simple graph with edge weights, we apply a GCN with self-loop to learn the graph structure and upgrade hyperedge embeddings $Y^{(l+1)}$. The output of HE2HE stage can be expressed as:

$$Y^{(l+1)} = \sigma \left( D_e^{-\frac{1}{2}} (H^T H + I) D_e^{-\frac{1}{2}} Y^{(l,1)} \theta_2^{(l)} \right) \quad (11)$$

where $Y^{(l,1)}$ represents the hyperedge embeddings that have aggregated node information. $\sigma$ denotes a non-linear activation function, $I$ represents the identity matrix, and $\theta_2^{(l)}$ is the trainable parameter matrix.

By adopting a similar approach as in the N2HE stage, using a mean aggregation method to aggregate hyperedge information can be represented as $D_v^{-1/2} H D_e^{-1/2} Y^{(l+1)}$. Adding the aggregated information to the original node information and introducing a hyperparameter $\beta$, the output of HE2N stage can be expressed as:

$$X^{(l+1)} = \beta D_v^{-\frac{1}{2}} H D_e^{-\frac{1}{2}} Y^{(l+1)} + X^{(l)} \quad (12)$$

introducing attention mechanism to compute the weight updates of the incidence matrix $H$ and optimizing the hypergraph structure using hyperedge outlier removal mechanism, the formulation of information propagation from hyperedges to nodes in the hypergraph can be expressed as:

$$e'_{ij} = (W'^T_Q x_i)^T (W'^T_K y_j) \quad (13)$$

$$\alpha'_{ij} = \frac{\exp(e'_{ij})}{\sum_{v_k \in N_{v_i}} \exp(e'_{ik})} \quad (14)$$

$$X^{(l+1)} = \sigma \left( \left( \beta D_v^{-\frac{1}{2}} hor(H^{Att'}) D_e^{-\frac{1}{2}} Y^{(l+1)} + X^{(l)} \right) \theta_3^{(l)} \right) \quad (15)$$

where $\beta$ is a hyperparameter, $W'_Q, W'_K, \theta_3^{(l)}$ are learnable parameter matrices, $H^{Att'}$ is the attention score incidence matrix computed in this stage, and $hor(\cdot)$ is the hyperedge outlier removal function.

Finally, the $l-th$ hypergraph convolution layer is constructed as:

$$Y^{(0)} = D_e^{-1} H^T X^{(0)} \quad (16)$$

$$Y^{(l)} = \sigma \left( \left( \alpha D_e^{-\frac{1}{2}} (H^{Att})^T D_v^{-\frac{1}{2}} X^{(l)} + Y^{(l)} \right) \theta_1^{(l)} \right) \quad (17)$$

$$Y^{(l+1)} = \sigma \left( D_e^{-\frac{1}{2}} (H^T H + I) D_e^{-\frac{1}{2}} Y^{(l)} \theta_2^{(l)} \right) \quad (18)$$

$$X^{(l+1)} = \sigma \left( \left( \beta D_v^{-\frac{1}{2}} H^{Att'} D_e^{-\frac{1}{2}} Y^{(l+1)} + X^{(l)} \right) \theta_3^{(l)} \right) \quad (19)$$

Considering the case that dataset lacks initial hyperedge features and the model only receives initial node features $X^{(0)}$ as input, it is necessary to calculate the average of the node features within each hyperedge to generate the corresponding hyperedge features $Y^{(0)}$ as equation (16). Moreover, if we remove the HE2HE stage and consider a convolution layer without self-loops, we can derive the following expression by multiplying equation (17) and (19):

$$X^{(l+1)} = \sigma \left( \beta \alpha D_v^{-\frac{1}{2}} H D_e^{-1} H^T D_v^{-\frac{1}{2}} X^{(l)} \theta^{(l)} \right) \quad (20)$$

which represents a hypergraph convolution layer as defined in HGNN [7]. Therefore, the hypergraph convolution in HGNN can be seen as a simplified form of HeIHNN. A hypergraph convolutional layer is shown as Algorithm 1.

---

**Algorithm 1** Hypergraph Convolution Algorithm

**Input**: feature matrix $X$,
    hypergraph graph $G = (V, E, H)$
**Parameter**: hyperparameters $\alpha, \beta$
**Output**: node embedding $X^L$,
    hyperedge embedding $Y^L$
1: **foreach** $v \in V$ **do** $x_v^0 \leftarrow X_v$
2: **foreach** $e \in E$ **do** $y_e^0 \leftarrow \sum_{v \in e} x_v^0$
3: **for** $l = 0\ to\ L - 1$ **do**
4:     **foreach** $e \in E$ **do**
5:         $Ne' \leftarrow Hor(X^l, Y^l, H)$
6:         $y_e^{l,'} \leftarrow \sum_{v \in Ne'} \alpha \cdot Mess(x_v^l) + y_e^l$
7:     **end**
8:     **foreach** $e \in E$ **do**
9:         $y_e^{l+1} \leftarrow \sum_{u \in Ne} y_u^{l,'} + y_e^l$
10:    **end**
11:    **foreach** $v \in V$ **do**
12:       $Nv' \leftarrow Hor(X^l, Y^l, H)$
13:       $x_v^{l+1} \leftarrow \sum_{e \ni Nv'} \beta \cdot Mess(y_e^{l+1}) + x_v^l$
14:    **end**
15: **end**
16: $X^L \leftarrow [x_{v1}^L, x_{v2}^L, \ldots, x_{vn}^L]^T$
17: $Y^L \leftarrow [y_{e1}^L, y_{e2}^L, \ldots, y_{em}^L]^T$
18: **return** $X^L, Y^L$

Table 2: Dataset statistics.

| Datasets | Cora | Citeseer | Pubmed | ModelNet40 | NTU2012 |
|---|---|---|---|---|---|
| $|V|$ | 2708 | 3312 | 19717 | 12311 | 2012 |
| $|E|$ | 1579 | 1079 | 7963 | 12311 | 2012 |
| max$|e|$ | 5 | 26 | 171 | 5 | 5 |
| features | 1433 | 3703 | 500 | 100 | 100 |
| classes | 7 | 6 | 3 | 67 | 40 |

Table 3: Results for the tested datasets: Mean accuracy (%) ± standard deviation.

|  | Cora | Citeseer | Pubmed | ModelNet40 | NTU2012 |
|---|---|---|---|---|---|
| MLP | 72.24±1.21 | 69.14±1.52 | 80.06±0.36 | 96.18±0.33 | 82.26±1.76 |
| GCN-CE | 72.75±1.13 | 67.43±1.38 | 79.38±0.56 | 88.87±0.28 | 80.42±0.79 |
| GAT-CE | 73.14±1.02 | 67.66±1.05 | 79.84±0.94 | 90.51±0.31 | 80.61±1.14 |
| HGNN | 75.92±1.51 | 69.18±0.67 | 80.21±0.31 | 96.90±0.05 | 84.45±1.24 |
| HyperGCN | 74.21±1.25 | 68.58±0.83 | 76.99±3.06 | 76.04±0.38 | 75.83±2.09 |
| HNHN | 73.14±1.25 | 68.91±0.53 | 81.14±0.26 | 97.36±0.43 | 85.27±1.62 |
| HCHA | 73.12±1.73 | 68.87±2.14 | 81.07±0.41 | 96.24±0.46 | 85.15±1.22 |
| AllDeepSet | 72.67±1.68 | 68.67±1.15 | 85.75±0.33 | 96.67±0.31 | 84.97±1.07 |
| AllSet-TF | 74.36±1.02 | 70.13±0.97 | 84.50±0.41 | 98.17±0.15 | 85.16±1.38 |
| HeIHNN | 78.91±0.67 | 68.37±0.55 | 88.67±0.26 | 97.69±0.43 | 85.43±1.17 |

## 5 EXPERIMENTS

### 5.1 Datasets

We conducted node classification tasks on three citation network datasets: Cora, Citeseer, Pubmed [29] and two visual object datasets: ModelNet40 [30], NTU2012 [31]. The dataset statistics are illustrated in Table 2: $|V|$ refers to the number of nodes in the dataset. $|E|$ refers to the number of hyperedges in the dataset. max$|e|$ refers to the maximum degree of a single hyperedge, which refers to the maximum number of nodes contained within a hyperedge.

Each data point in these citation network datasets is represented by a bag-of-words representation of documents, and each 3D object in visual object datasets is represented by its extracted features using MVCNN [33] and GVCNN [34]. To construct the hypergraph structure on these datasets, we followed the approach proposed in [14]. For the citation network datasets, we constructed hyperedges by considering the nodes and their neighbors within the graph structure. And for the visual object datasets, we constructed hyperedges based on the similarity of node features. For each dataset, 80% of the data points were randomly selected for training, while the remaining 20% were used for testing.

### 5.2 Experimental Setups

To reduce computational complexity, HeIHNN used in our experiments only included 1-step hyperedge propagation. We utilized the softmax function to predict labels for performing node classification tasks. In the training phase, we employed the cross-entropy loss on the training data to backpropagate and update the model parameters. In the testing phase, we predicted the labels for the test data and evaluated the performance.

We compare our HeIHNN with several baseline models, MLP, CE+GCN [1], CE+GAT [2], HGNN [7], HyperGCN [12], HNHN [36], HCHA [11], AllDeepSet and AllSetTransformer [20]. Here CE means clique expansion on hypergraph.

For baseline implementations, we use the source code and set the optimal parameters provided in the corresponding papers. For the proposed HeIHNN, the learning rate and weight decay are set to 0.001 and 0.0005 respectively. The dropout is employed to avoid overfitting with drop rate $p = 0.5$.

### 5.3 Experimental Results

The experimental results on each dataset and the comparison results are shown in Table 3: the cells shaded in dark blue represent the best results, while the cells shaded in light blue indicate the second-best results. We used accuracy (micro-F1 score) and their standard deviations as evaluation metrics. The results indicate that HeIHNN achieves the best or close-to-best performance compared to the best methods on both types of datasets. On the citation network datasets, Cora and Pubmed, our model outperforms existing models with a significant improvement of over 3%. On Citeseer, our model performed slightly worse with an accuracy that was 2% lower than the best model. On ModelNet40, our model achieves around 0.6% improvement compared to HGNN, slightly lower than the state-of-the-art model AllSet-Transformer. On NTU2012, our model achieves approximately 1% improvement compared to HGNN.

The enhanced performance of HeIHNN can be attributed to the HE2HE propagation stage during the convolution process. as indicated by the convolutional layer formula, HeIHNN captures additional insights by incorporating the inter-hyperedge interaction information. Secondly, the optimization of the hypergraph structure through the HOR mechanism further network models, HeIHNN is expected to perform better on datasets with more complex and abundant collective interactions.

Table 4: Ablation study results for the tested datasets: Mean accuracy (%) ± standard deviation.

|  | Cora | Citeseer | Pubmed | ModelNet40 | NTU2012 |
|---|---|---|---|---|---|
| Neither | 78.43±0.58 | 67.65±0.37 | 83.35±0.38 | 97.14±0.98 | 84.97±0.49 |
| Only-S1 | 78.40±1.02 | 68.16±0.45 | 88.62±0.33 | 97.69±0.43 | 85.43±0.17 |
| Only-S3 | 78.91±0.77 | 67.46±0.49 | 88.46±0.18 | 97.40±0.58 | 84.89±0.23 |
| Both | 78.77±0.67 | 68.37±0.55 | 88.67±0.26 | 97.52±0.46 | 85.34±0.13 |

Table 5: Robustness analysis results for the tested datasets.

|  | Cora | | Citeseer | | Pubmed | | ModelNet40 | | NTU2012 | |
|---|---|---|---|---|---|---|---|---|---|---|
|  | ptAcc | Dec | ptAcc | Dec | ptAcc | Dec | ptAcc | Dec | ptAcc | Dec |
| GCN-CE | 65.00 | 7.52 | 58.67 | 8.76 | 72.93 | 6.45 | 83.10 | 5.77 | 70.68 | 9.74 |
| HyperGCN | 66.48 | 7.73 | 60.02 | 8.56 | 69.93 | 7.06 | 72.62 | 3.42 | 70.15 | 5.68 |
| HGNN | 68.51 | 7.41 | 61.23 | 7.95 | 73.43 | 6.78 | 91.27 | 5.63 | 75.35 | 9.10 |
| HCHA | 64.99 | 8.13 | 59.75 | 9.12 | 75.18 | 5.89 | 94.14 | 2.07 | 78.42 | 6.73 |
| HeIHNN | 72.73 | 6.16 | 60.83 | 7.54 | 82.69 | 5.98 | 95.94 | 1.75 | 81.03 | 4.31 |

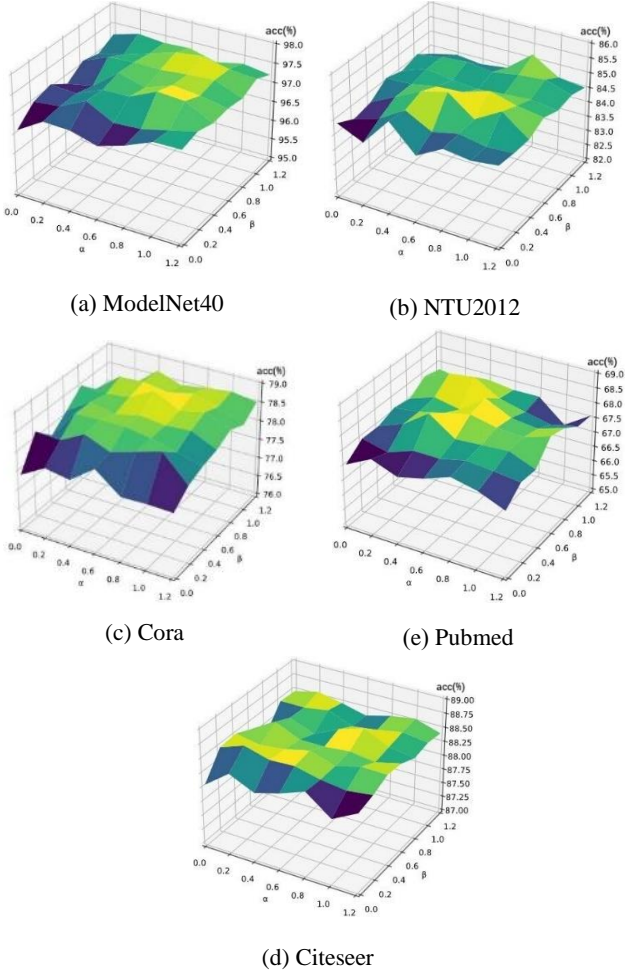

(a) ModelNet40  (b) NTU2012
(c) Cora  (e) Pubmed
(d) Citeseer

**Figure 3: Parameter sensitivity of the hyper-parameters.**

### 5.4 Ablation Study

To demonstrate the effectiveness of the proposed HOR mechanism on hypergraph structure optimization and its influence on different stages, we designed the following ablation experiment. There are two stages in the propagation process where HOR can be applied: N2HE and HE2N. We conducted 10 repeated experiments and compared the results on the datasets, sequentially applying the following approaches: neither stage use HOR, only in N2HE, only in HE2N, and both.

The ablation study results are shown in Table 4, the cells shaded in blue refers to the best results. Neither indicates that neither stage uses HOR. Only-S1 implies that HOR was applied only in the N2HE stage. Only-S3 indicates that HOR was used only in the HE2N stage. Both signifies that HOR was employed in both stages. It can be seen that the models utilizing HOR mechanism outperform the models without it. This indicates the effectiveness of the mechanism in optimizing the structure of the hypergraph. HOR helps remove outliers in the hyperedges before the information propagation. In the citation network datasets, Cora and Citeseer, the improvement achieved by the HOR mechanism appears to be relatively weak. This can be attributed to two main factors. Firstly, when expanding the subgraphs' degree up to 3, the underlying graph structure remains largely unchanged. As a result, the generated hypergraph structure is likely to closely resemble the original graph structure [12]. Secondly, in these datasets, most of the edges represent reliable and meaningful relationships. This characteristic makes it less likely for outliers to exist within hyperedges. Additionally, determining the optimal stage to apply the mechanism remains a question.

### 5.5 Parameter Analysis

In this section, we investigate the effect of two hyper-parameters on our model: the coefficients ($\alpha$ and $\beta$) of propagated information. We vary the values of $\alpha$ and $\beta$ and show the results.

contributes to the improved performance. An optimized hypergraph structure can better represent data high-order correlations. Additionally, compared to other hypergraph neural

Specifically, α and β both is in the range from {0.0, 0.2, 0.4, 0.6, 0.8, 1.0, 1.2}. The results in terms of accuracy (micro-F1 score) are visualized in Figures 3.

As can be seen from the visualization, HeIHNN has poor performance when $\alpha = 0.0$ or $\beta = 0.0$. This is because setting the parameters to 0 means that the corresponding information propagation is not performed. On the contrary, when the parameters have large values, few of the original features are retained, leading to a decline in the model's performance. Finally, it can be seen from the visualization that HeIHNN is less sensitive to the two hyperparameters in general and achieves high performance over a relatively wide range of values.

## 5.6 Robustness Analysis

In this section, we focus on the ability of model to handle noise and outlier information in the dataset. We employ the Projected Gradient Descent (PGD) [38] method with a parameter setting of $\varepsilon = 0.002$ to introduce perturbations into the node features of the dataset, generating a perturbed feature matrix. We replace the feature matrix of the test set with the perturbed feature matrix and evaluate the trained model. The accuracy of HeIHNN and the baseline models on the perturbed test set, as well as the drop in accuracy compared to the original test set, are shown in Table 5. The ptAcc refers to accuracy (%) on perturbed datasets, Dec refers to decrease in accuracy (%) compared to clear datasets. The cells shaded in blue refers to the best results.

The experimental results demonstrate that HeIHNN performs admirably on the perturbed dataset and exhibits a lower decrease in accuracy compared to the baseline across multiple datasets. This suggests that HeIHNN possesses enhanced capabilities in effectively handling noise and outlier information.

## 6 CONCLUSION

In this paper, we proposed a novel hypergraph neural net-work, HeIHNN. HeIHNN comprises three stages of hyper-graph convolution, designed to capture high-order information from hyperedge interactions. During the information propagation stages between hyperedges and nodes, Outlier Removal Mechanism (HOR) is employed, dynamically adjusting the hyperedge structure during the training process. Our experiments on citation networks and 3D visual object datasets demonstrate the competitive performance of HeIHNN. Due to the characteristics of HeIHNN, it can achieve remarkable results in scenarios with a large number of hyperedge interactions. Furthermore, the ability to handle outliers brought by the Outlier Removal Mechanism (HOR) allow HeIHNN to perform well in scenarios with noisy data.